\documentclass[numbers]{assets/matterlab}

\usepackage{microtype}
\usepackage{graphicx}
\usepackage{booktabs}
\usepackage{float}
\usepackage{xurl}
\usepackage{hyperref}
\usepackage{titlesec}

\usepackage[nohyperlinks]{acronym}
\AtBeginDocument{
\acrodef{ai}[AI]{artificial intelligence}
\acrodef{api}[API]{application programming interface}
\acrodef{cad}[CAD]{computer-aided design}
\acrodef{dmta}[DMTA]{design--make--test--analyse}
\acrodef{sdl}[SDL]{self-driving laboratory}
\acrodefplural{sdl}[SDLs]{self-driving laboratories}
\acrodef{sbl}[SBL]{self-building laboratory}
\acrodefplural{sbl}[SBLs]{self-building laboratories}
\acrodef{si}[SI]{supporting information}
\acrodef{urdf}[URDF]{Unified Robot Description Format}
\acrodef{llm}[LLM]{large language model}
\acrodef{usd}[USD]{Universal Scene Description}
\acrodef{nmr}[NMR]{nuclear magnetic resonance}
\acrodef{thf}[THF]{tetrahydrofuran}
\acrodef{tbadt}[TBADT]{tetrabutylammonium decatungstate}
\acrodef{hplc}[HPLC]{high-performance liquid chromatography}
}

\usepackage{amsmath}
\usepackage{amssymb}
\usepackage{mathtools}
\usepackage{amsthm}
\usepackage{bm}

\usepackage[noabbrev,nameinlink]{cleveref}

\usepackage{amsmath,amsfonts,bm}

\def\eqref#1{equation~\ref{#1}}

\def\1{\bm{1}}

\DeclareMathAlphabet{\mathsfit}{\encodingdefault}{\sfdefault}{m}{sl}
\SetMathAlphabet{\mathsfit}{bold}{\encodingdefault}{\sfdefault}{bx}{n}

\usepackage{orcidlink}
\usepackage{comment}
\usepackage[version=4]{mhchem}
\usepackage{longtable}
\usepackage{array}
\renewcommand{\arraystretch}{1.7}
\usepackage{multirow}           
\usepackage{amsfonts}           
\usepackage{nicefrac}
\usepackage{duckuments}         
\usepackage{thmtools,thm-restate}
\usepackage{enumitem}
\usepackage{xcolor,colortbl}
\usepackage{tikz}
\usepackage{tikz-cd}
\usepackage{caption}
\usepackage{subcaption}
\usepackage{listings}
\usepackage{gensymb}
\usepackage{textcomp} 
\usepackage[inkscapelatex=false]{svg}

\newcommand{\addressCHEM}{Department of Chemistry, University of Toronto,  80 St. George St., Toronto, ON M5S 3H6, Canada}
\newcommand{\addressAC}{Acceleration Consortium, 700 University Ave., Toronto, ON M7A 2S4, Canada}
\newcommand{\addressCS}{Department of Computer Science, University of Toronto, 40 St George St., Toronto, ON M5S 2E4, Canada}
\newcommand{\addressVECTOR}{Vector Institute for Artificial Intelligence, W1140-108 College St., Schwartz Reisman Innovation Campus, Toronto, ON M5G 0C6, Canada}
\newcommand{\addressMSE}{Department of Materials Science \& Engineering, University of Toronto, 184 College St., Toronto, ON M5S 3E4, Canada}
\newcommand{\addressCHEMENG}{Department of Chemical Engineering \& Applied Chemistry, University of Toronto, 200 College St., Toronto, ON M5S 3E5, Canada}
\newcommand{\addressMEDICALSCI}{Institute of Medical Science, 1 King's College Circle, Medical Sciences Building, Room 2374, Toronto, ON M5S 1A8, Canada}
\newcommand{\addressCIFAR}{Canadian Institute for Advanced Research (CIFAR), 661 University Ave., Toronto,
ON M5G 1M1, Canada}
\newcommand{\addressNVIDIA}{NVIDIA, 431 King St W \#6th, Toronto, ON M5V 1K4, Canada}

\newcommand{\acknowAC}{This research is part of the University of Toronto’s Acceleration Consortium, which receives funding from the CFREF-2022-00042 Canada First Research Excellence Fund. }

\newcommand{\acknowDARPA}{This work was supported by the Defense Advanced Research Projects Agency (DARPA) under Agreement No.~HR0011262E022.}

\providecommand{\doi}[1]{\href{https://doi.org/#1}{doi:#1}}

\usepackage{soul, color}
\newcommand{\Note}[1]{}
\renewcommand{\Note}[1]{#1}  

\title{Self-Building Labs}

\author[\dagger,1,3,\orcidlink{0000-0002-1246-1993}]{Jiaru~Bai}
\author[\dagger,2,3,\orcidlink{0009-0006-5419-4391}]{Yuchi~Zhao}
\author[\ddagger,1,4,\orcidlink{0000-0002-0176-3929}]{Han~Hao}
\author[\ddagger,5,\orcidlink{0009-0009-1154-7005}]{Elia~Savino}
\author[2,3,4,\orcidlink{0000-0001-7255-7658}]{Miroslav~Bogdanovic}
\author[5,\orcidlink{0009-0009-4660-2019}]{Derk~H.W.~ten~Klooster}
\author[5,\orcidlink{0009-0001-9509-5414}]{Simone~Pilon}
\author[5,\orcidlink{0000-0002-3107-6927}]{Timothy~Noël}
\author[1,4,\orcidlink{0000-0001-8636-7919}]{Yang~Cao}
\author[1,2,4,*,\orcidlink{0000-0002-8446-7956}]{Varinia~Bernales}
\author[1,2,3,4,6,7,8,9,10,*,\orcidlink{0000-0002-8277-4434}]{Al\'an~Aspuru-Guzik}

\affiliation[1]{\addressCHEM}
\affiliation[2]{\addressCS}
\affiliation[3]{\addressVECTOR}
\affiliation[4]{\addressAC}
\affiliation[5]{Flow Chemistry Group, van’t Hoff Institute for Molecular Sciences (HIMS), Universiteit van Amsterdam (UvA), 1098 XH, Amsterdam, The Netherlands}
\affiliation[6]{\addressMSE}
\affiliation[7]{\addressCHEMENG}
\affiliation[8]{\addressMEDICALSCI}
\affiliation[9]{\addressCIFAR}
\affiliation[10]{\addressNVIDIA}

\contribution[\dagger]{These authors contributed equally}
\contribution[\ddagger]{these authors also contributed equally.}

\abstract{
\Acp{sdl} accelerate scientific discovery by automating experimental execution and optimization. AI agents can increasingly propose new experiments, but existing \acp{sdl} can only run the procedures they were built for. Adapting them to a new experiment still requires extensive manual engineering, which makes laboratory flexibility the bottleneck for scaling discovery. We introduce the \ac{sbl}, which designs, configures, and operates its own laboratory setup given a high-level scientific goal. \ac{sbl} starts from whatever already exists, from an empty room to a working \ac{sdl}, and builds the missing capabilities: the lab layout, its digital twin, robot skills, and experimental workflows. We demonstrate the system working across different robots and setups, scaling from a single workstation to laboratories with several robots working together. We validate \ac{sbl}'s end-to-end capabilities through a real-world quantum-dot synthesis experiment. By shifting human researchers from manually configuring and programming device automation to specifying scientific objectives, \ac{sbl} makes scientific automation more accessible, flexible, and scalable.

}

\metadata[Website]{\href{https://self-building-labs.github.io/}{\texttt{self-building-labs.github.io}}}
\date{\today}
\correspondence{\email{alan@aspuru.com}, \email{varinia@bernales.org}}

\begin{document}

\maketitle



\newpage
\section{Introduction}

\Acfp{sdl} accelerate scientific research by automating experiments through the \ac{dmta} cycle, in which experimental results inform subsequent decisions~\cite{aspuru2018materials,roch2018chemos,steiner2019organic,coley2019robotic,macleod2020self,burger2020mobile,gromski2020universal,sim2024chemos,Tom2024sdl,volk2023alphaflow, slattery2024robochem,bai2024dynamic}.
These approaches have been used to design biological systems, synthesize solid-state materials, optimize nanocrystals~\cite{hamedirad2019towards,Szymanski2023,Xu2025}, to name a few.
However, the autonomy of an \ac{sdl} is typically bounded by the equipment and operational capabilities established during its development~\cite{Canty2026}.

This boundary becomes a bottleneck as scientific research evolves.
New research questions may require experimental procedures or instrumentation that were not anticipated when the platform was configured.
Extending these capabilities typically requires specialists to integrate hardware and develop the software needed to execute new procedures~\cite{bai2022platform,bai2024dynamic,rahmanian2022enabling}.
Such modifications are time-consuming~\cite{MacLeod2023howSDL} and often depend on tacit knowledge that is difficult to preserve or transfer~\cite{christensen2021automation}.
Changes in equipment placement or laboratory layout can further compromise the reliability of existing workflows.
Modular platforms and robots that reconfigure equipment can adapt workflows using existing hardware and software~\cite{bedard2018reconfigurable,coley2019robotic,pilon2026robochemflex}.
Capabilities beyond those already available still require human development.

\begin{figure}[!ht]
    \centering
\includegraphics[width=\linewidth]{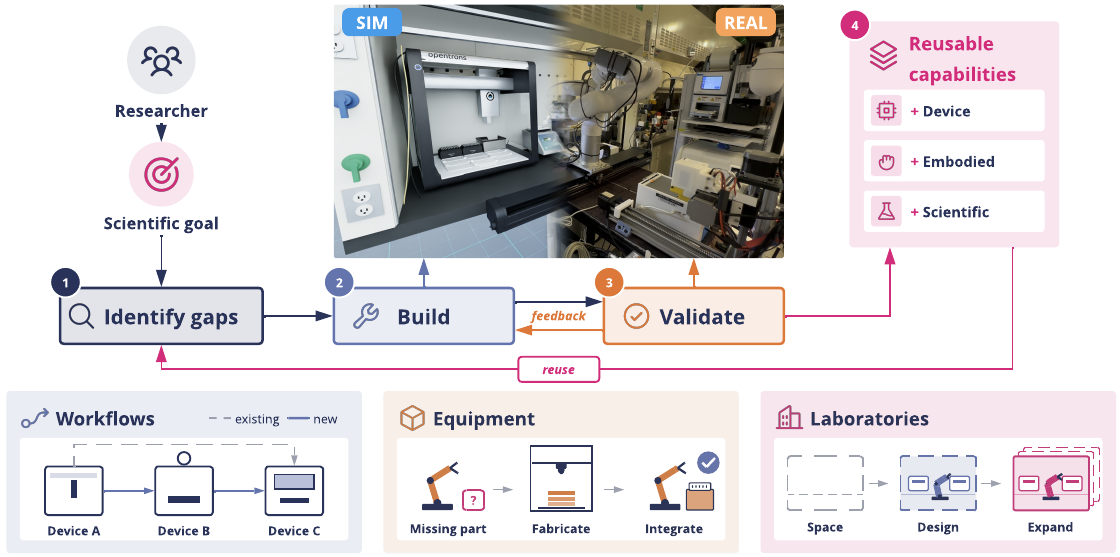}
    \caption{
    \textbf{Concept of a self-building laboratory (SBL)}. Scientific goals guide the identification of capability gaps, development in a digital twin, and validation in the physical laboratory. Feedback and reuse support continued capability development across workflows, equipment, and entire laboratories. The image combines SDL2 at the Acceleration Consortium, University of Toronto, Canada, with its digital twin built by the SBL and rendered in Isaac Sim.
    }
    \label{fig:sbl}
\end{figure}

Recent advances provide an opportunity to address this limitation.
\Ac{llm}-based agents have demonstrated the ability to plan experiments, generate instrument control code, and interact with laboratory hardware~\cite{Zou2025ElAgente,Yoshikawa2023llm_chem_robots,pagel2026chemputation,Boiko2023,chemcrow,Ding2025SciToolAgent,grafico,laagenteoptima}, while physics-based simulation enables developing digital twins of laboratory environments~\cite{Darvish2026matterix,rihm2024transforming}.
Advances in robotics, particularly mobile manipulation, allow experiments to draw on instruments distributed across a laboratory by transferring samples between instruments~\cite{burger2020mobile,Darvish2024organa,Dai2024Autonomous, yoshikawa2023chemistrylabautomationconstrained,Longley2026RobInHood}.
Language models can select existing robot skills based on their likelihood of success~\cite{Ahn2022SayCan} and generate control programs that use sensory feedback~\cite{Liang2022CodeAsPolicies}.
Using robots as tools for general-purpose agents has also been discussed as a broader direction for robotics~\cite{Isola2026RobotUseAgents}.
Recent studies test GPT-6-Astra for robot control in simulation~\cite{Zhang2026RoboDojo,Galbot2026Astra,Yang2026Astranomy}.
Experiments and technical reports also show control of real robots, with some systems reusing experience or using a digital twin to guide actions~\cite{He2026Astra,Huh2026HomeBody,Robocurve2026Astra}.
These advances support flexible experimental execution, but adapting a laboratory to new scientific requirements still requires substantial time and expertise~\cite{Cooper2025}.

Here, we demonstrate a \textbf{\acf{sbl}} that incorporates the identification, development, and validation of task-required operational capabilities into the autonomous scientific process (\Cref{fig:sbl}).
Researchers define the scientific objectives that guide this process.
\ac{sbl} self-identifies gaps between these objectives and its available capabilities, then develops or extends the interfaces, skills, and workflows needed to address them.
The system uses a digital twin to develop candidate operations and check them against task constraints before refining and validating them through physical interaction.
Successful procedures are retained as reusable skills for subsequent tasks.
\Cref{tab:laboratory-comparison} compares how experimental capabilities are developed and used across laboratory paradigms.

We examine \acp{sbl} capability development through digital twin construction, operational skill development, experimental workflow building, and transport prediction.
In a quantum-dot synthesis study, \ac{sbl} constructs a task-relevant digital twin and develops interfaces and feedback-guided skills that connect liquid handling, robotic plate transfer and optical measurement.
On the existing RoboChem-Flex platform~\cite{slattery2024robochem,pilon2026robochemflex}, an agent establishes operating procedures for flow photochemistry and inline analysis under expert supervision.
Separate transport studies on that platform develop and test a calibrated model for predicting downstream liquid and gas arrivals under different pump programmes.
In simulation, the system also coordinates different robotic configurations and interconnected workstations.
By incorporating capability development into the autonomous scientific process, \acp{sbl} could reduce the engineering effort required to adapt laboratory automation to new scientific objectives and make it more accessible to researchers without specialized automation expertise.

\begin{table}[!ht]
    \centering
    \caption{Comparison of laboratory paradigms.}
    \label{tab:laboratory-comparison}
    \small
    \setlength{\tabcolsep}{4pt}
    \renewcommand{\arraystretch}{1.2}
    \begin{tabular}{@{}>{\raggedright\arraybackslash}p{\dimexpr0.19\linewidth-2\tabcolsep\relax}>{\raggedright\arraybackslash}p{\dimexpr0.24\linewidth-2\tabcolsep\relax}>{\raggedright\arraybackslash}p{\dimexpr0.27\linewidth-2\tabcolsep\relax}>{\raggedright\arraybackslash}p{\dimexpr0.30\linewidth-2\tabcolsep\relax}@{}}
        \toprule
        \textbf{Aspect} & \textbf{Traditional lab} & \textbf{Self-driving lab} & \textbf{Self-building lab (this work)} \\
        \midrule
        Experiment execution & Researcher-led procedures & Automated workflows using integrated capabilities & Workflows using capabilities developed as needed \\
        New capabilities & Developed by researchers & Typically engineered by specialists & Developed and validated by agents \\
        Knowledge retention & Written procedures and researcher experience & Protocols and script libraries tailored for one setup & Reusable skills across laboratories \\
        \bottomrule
    \end{tabular}
\end{table}

\section{Results}

\subsection{Overview}

\begin{figure}[!ht]
    \centering
    \includegraphics[width=\linewidth]{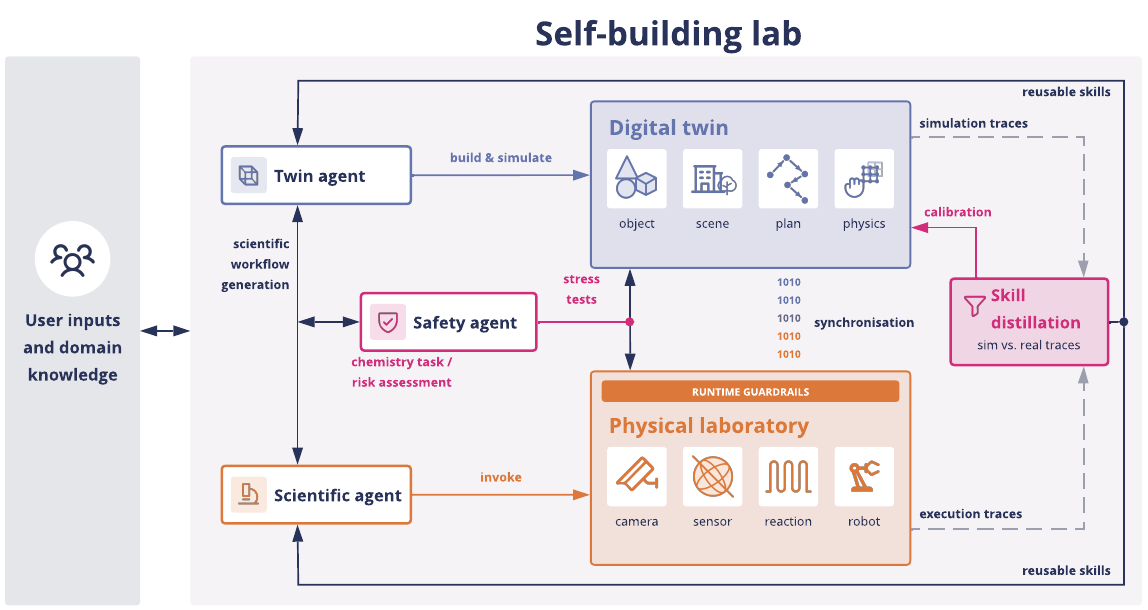}
    \caption{\textbf{Architecture of the self-building laboratory.}
    The \textbf{scientific agent} translates user goals into experimental plans, while the \textbf{twin agent} constructs a \textbf{digital twin}, develops task-relevant skills, and validates workflows in simulation.
    The \textbf{safety agent} uses the twin for scenario analysis and generates \textbf{runtime guardrails} through which all physical laboratory commands pass.
    The scientific agent invokes the \textbf{physical laboratory} through this layer.
    Physical state feedback supports twin calibration and subsequent simulation, while simulation and physical execution traces support \textbf{skill distillation} and retain experience as agent memory.
    Reusable skills and memories inform subsequent tasks.
    }
    \label{fig:overview}
\end{figure}

\begin{table}[!ht]
    \centering
    \caption{Starting points for a self-building laboratory. The same process applies in each case; only the required set of capabilities differs.}
    \label{tab:starting-points}
    \small
    \setlength{\tabcolsep}{4pt}
    \renewcommand{\arraystretch}{1.2}
    \begin{tabular}{@{}>{\raggedright\arraybackslash}p{\dimexpr0.32\linewidth-2\tabcolsep\relax}>{\raggedright\arraybackslash}p{\dimexpr0.26\linewidth-2\tabcolsep\relax}>{\raggedright\arraybackslash}p{\dimexpr0.42\linewidth-2\tabcolsep\relax}@{}}
        \toprule
        \textbf{Starting point} & \textbf{Already exists} & \textbf{\ac{sbl} builds} \\
        \midrule
        Empty laboratory & N/A & Layout, digital twin, skills, and workflows \\
        Existing manual laboratory & Equipment & Digital twin, skills, and workflows \\
        Automated laboratory requiring reconfiguration & Equipment and some skills & New layout, updated digital twin, and adapted skills \\
        \ac{sdl} with a new protocol & Equipment and skills & New workflows and any missing skills \\
        \bottomrule
    \end{tabular}
\end{table}

\ac{sbl} takes a user's scientific goal together with a description of the current laboratory: photographs of the space, device identifiers, and any existing automation interfaces (\Cref{fig:overview}).
The starting point can range from an empty room to a working \ac{sdl}; in each case, \ac{sbl} builds the capabilities missing for the goal: the lab layout, its digital twin, robot skills, and experimental workflows (\Cref{tab:starting-points}).
It builds a digital twin of the laboratory, from individual device models to the layout of the whole space, using robot-mounted cameras where available to gather additional observations (\Cref{building-digital-twin}).
\ac{sbl} identifies the capabilities the desired workflow requires and acquires the missing skills using the digital twin and, with user approval, trial-and-error in the real world (\Cref{building-skills}).
This allows it to go from a user's scientific goal to safely executing the desired workflow in the real world, with minimal human intervention (\Cref{building-workflows}).
The RoboChem-Flex study examines how an agent develops operating procedures and predictive transport capabilities on an existing automated platform (\Cref{sec:robochem}).
The same process extends to designing laboratory layouts and coordinating several robots across workstations (\Cref{building-labs}).

\subsection{Building digital twin}
\label{building-digital-twin}

\begin{figure}[p]
    \centering
    \includegraphics[width=\linewidth]{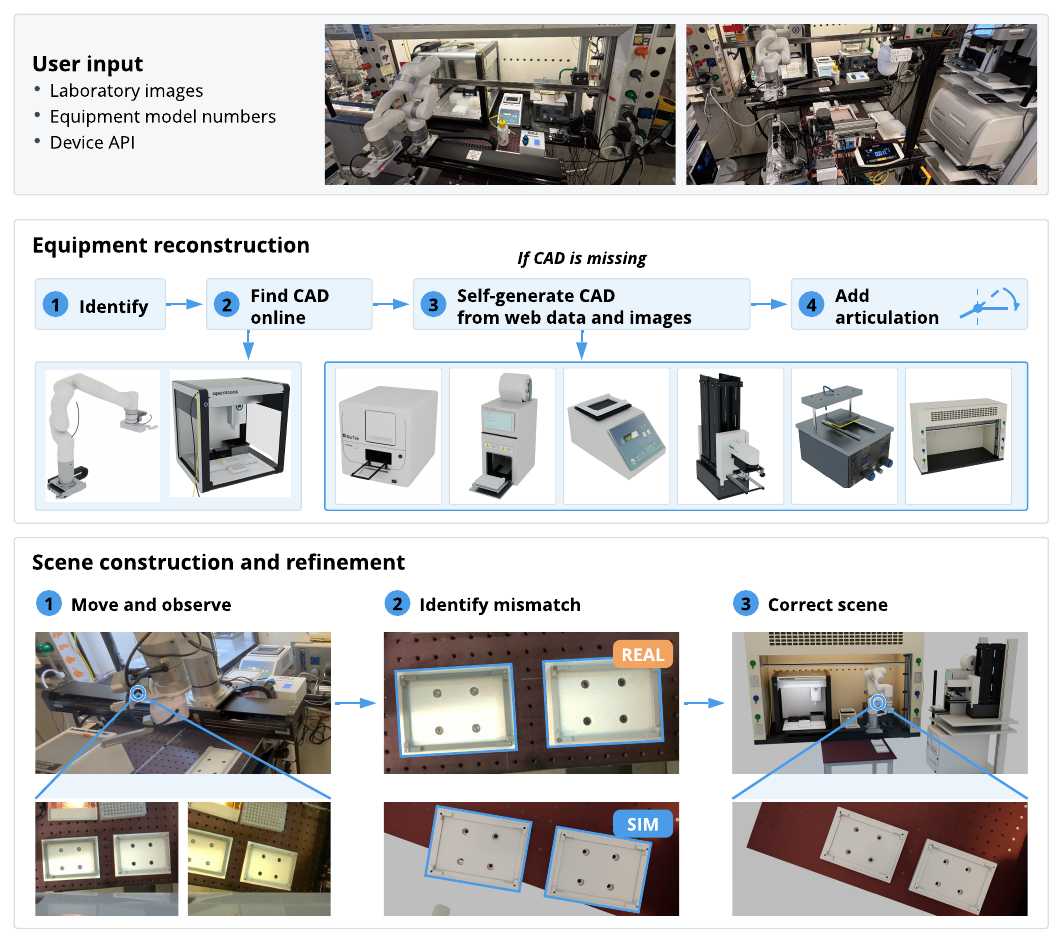}
    \caption{\textbf{Building a digital twin of the laboratory.}
    (a) Starting from laboratory images and equipment model numbers when available, the twin agent identifies individual devices, searches product documentation, and retrieves or constructs equipment models with task-relevant geometry, articulation, and control interfaces.
    (b) These models are assembled into an initial simulation scene and refined using colour and depth observations acquired from multiple robot poses.
    The agent compares observed and rendered views and iteratively updates equipment poses to align the scene with the physical laboratory.
    }
    \label{fig:twin}
\end{figure}

\Cref{fig:twin} shows how the twin agent reconstructs individual equipment and assembles these models into a calibrated laboratory scene.
The user provides laboratory photographs, equipment model numbers when available, and device \ac{api} endpoints.
From these inputs, the agent builds a task-oriented representation of equipment identity, geometry, articulation, spatial arrangement, and control interfaces.

Equipment reconstruction begins with identifying the individual devices from the photographs and any supplied model numbers.
For each device, the agent first searches for existing \ac{cad} models and \ac{urdf} files. If these are unavailable, the agent reconstructs an approximate 3D model using manufacturer documentation, technical specifications, and reference images. The agent first estimates the device dimensions using available measurements and known reference objects, then breaks down the device into individual components, such as frames, housings, doors, and moving mechanisms. Using Python, the agent constructs these components from basic geometric primitives and custom meshes. The agent then iteratively refines the generated models by comparing rendered views with reference images and adjusting geometry, proportions, and relative positions. Once the geometry is established, the agent adds articulations for movable components by defining their joints, motion axes, and limits. This process produces simulation-ready equipment models without requiring manual CAD modelling. Missing or uncertain parameters are recorded for further refinement.

The agent then assembles the equipment models into an initial scene based on the observed laboratory layout.
To refine this arrangement, the agent plans camera poses and acquires colour and depth observations from each pose using its wrist-mounted camera.
The agent compares observed and rendered views and iteratively updates equipment positions and orientations.
This procedure reduces equipment-pose discrepancies and establishes the robot-to-device poses needed for motion planning.

To support operations within the reconstructed scene, the agent implements simulated devices with the same \ac{api} interfaces as their physical counterparts but different endpoints.
Selecting the target endpoint determines whether an operation runs in the twin or the physical laboratory.
A monitoring system polls physical device status through the \acp{api} and updates the corresponding states in the twin.
Using the calibrated scene, the agent plans robot motion in the twin and sends the resulting trajectories to the physical execution endpoint to control the robot.
The twin remains a living replica of the physical laboratory, continually updated as new observations become available.

\subsection{Building laboratory skills}
\label{building-skills}

\begin{figure}[!ht]
    \centering
    \includegraphics[width=\linewidth]{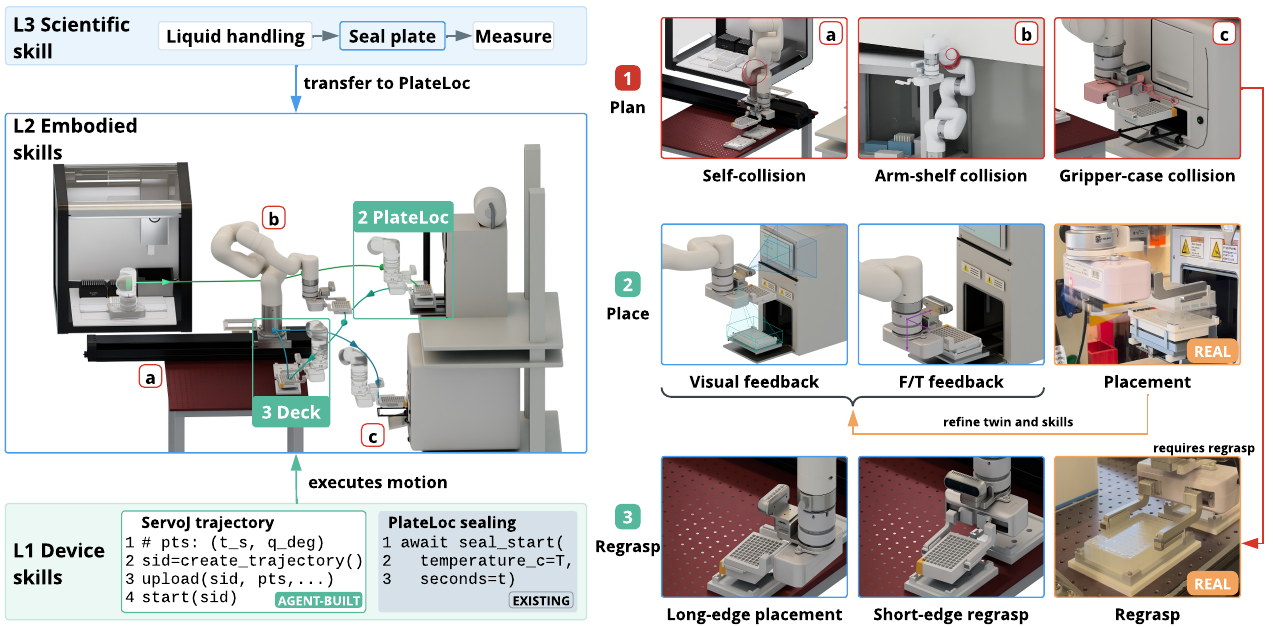}
    \caption{\textbf{Scientific tasks drive laboratory skill development.} A plate-sealing task drives the development and integration of embodied and device skills. Digital-twin planning and feedback from physical execution guide refinement of the twin and skills, connecting scientific requirements to executable laboratory capabilities.}
    \label{fig:skills}
\end{figure}

\Cref{fig:skills} illustrates laboratory skills at three levels: device execution interfaces, embodied operations, and scientific procedures and interpretation.
Scientific requirements create demand for embodied skills, which in turn depend on the capabilities exposed by device interfaces.
As these requirements arise, the agent develops missing skills and interfaces, building the operational capabilities needed to complete the task.

Physical skills that enable the robot to perform experimental tasks are developed by the agent in simulation and on the real robot. In simulation, the agent maps the required experimental operations onto available equipment and explores candidate action sequences through trial and error. Specifically, using object states, the agent automatically defines grasping and placement poses, then uses a motion planner~\cite{curobo_icra23} to generate collision-free trajectories through the cluttered laboratory workspace. It composes these motion segments into an experimental workflow. 

Directly transferring a workflow from simulation to the real world is limited by calibration and modelling errors.
Transport and coarse localization can often be directly deployed.
In contrast, precise interactions, like grasping and insertion that require millimetre-scale precision, are more sensitive to these errors.
There, online exploration is needed.
The agent can autonomously use local sensory feedback to refine those interactions on the real robot.
During exploration, it selects actions based on visual observations, depth measurements, robot state, and force/torque feedback, then examines the resulting sensory responses to identify successful strategies.
Through this process, it learns to interpret feedback and use it to guide subsequent actions.
Placement into a constrained fixture, for example, can involve visual alignment, guarded descent, contact detection, lateral adjustment, and resumed insertion after the measured load decreases.

A particularly challenging example is learning to operate an \ac{hplc} instrument originally designed for manual operation.
Loading a well plate into its holder requires sub-millimetre positioning accuracy, while opening and closing the drawer involve sustained contact and constrained mechanical motion (\Cref{fig:si-hplc}).
The robot must engage a spring-loaded locking mechanism and pull the drawer along its axis.
Before planning subsequent actions, it must determine how far the drawer has opened and where the holder is now located.
Limited camera visibility, narrow clearances, and the curved inner surface of the drawer further constrain the choice of contact points and manipulation trajectories.
The robot must also disengage without applying excessive force or torque.
To address these challenges, the agent modelled the drawer's articulation in the digital twin, explored alternative manipulation strategies, and refined the model and action sequence based on observations from physical trials.

The agent can complete a range of tasks using this approach through repeated cycles of reasoning, code generation, physical action, and observation.
However, these cycles were too slow to meet the requirements of a chemical workflow.
To reuse the experience gained during exploration, the system distills successful action sequences and their associated feedback into executable skills.
Once an agent establishes a repeatable procedure, it encodes it as a Python script containing deterministic actions, sensor-based conditions, and recovery branches, then tests it through repeated physical execution.
The resulting skill can then run without agent involvement during nominal operation.
Agentic reasoning resumes only when execution departs from the conditions encoded with the skill.

\paragraph{\textbf{Safety Guardrail.}}
Safety is a critical consideration when deploying autonomous agents in physical laboratories, particularly in environments shared with human researchers~\cite{leong2025steering,Kang2026seguro,Chen2026autonomyHarness}. To address this challenge, SBL incorporates safety mechanisms at both the simulation and physical execution levels. Specifically, beyond collision-free motion planning, SBL enables safety discovery by leveraging AI agents to autonomously propose and evaluate potentially hazardous scenarios in simulation. Through large-scale exploration of possible failure conditions, we demonstrate that SBL identifies potential risks and generates candidate safety guidelines that can be incorporated into subsequent workflow planning.
These findings also provide a foundation for building a shared knowledge base of laboratory hazards and safety practices. During physical execution, SBL implements multiple layers of safety guardrails. A separate control interface allows human operators to stop the entire experimental workflow at any time. Additionally, SBL enforces speed limits on robot trajectories and continuously monitors onboard force sensors to detect unexpected contact and trigger protective stops. Together, these capabilities extend SBL beyond conventional safety enforcement toward proactive hazard discovery and risk-aware autonomous laboratory operation.

\subsection{Building experimental workflow}
\label{building-workflows}

\begin{figure}[htbp]
    \centering
    \includegraphics[
        width=0.9\linewidth
    ]{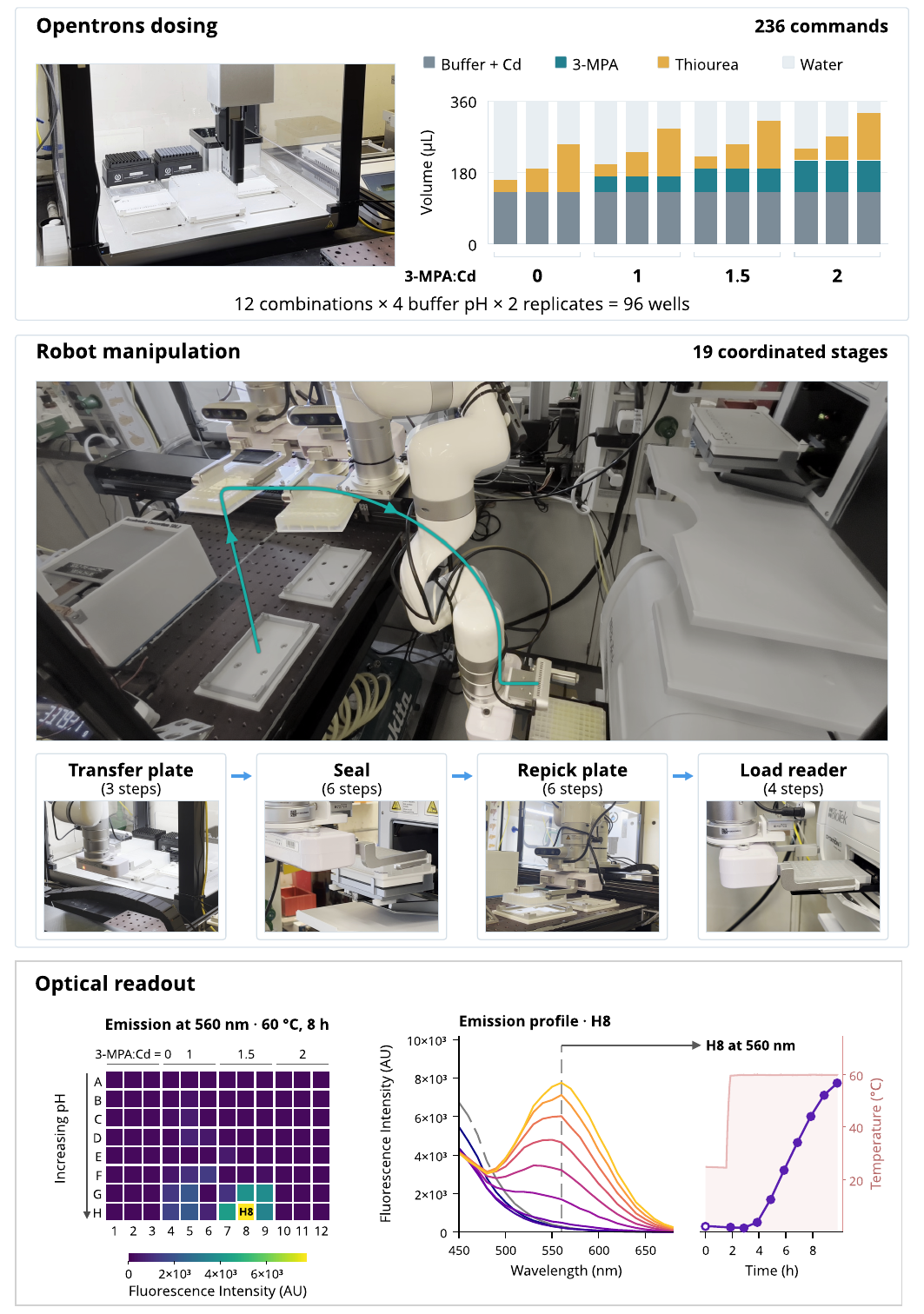}
    \caption{\textbf{\ac{sbl}-coordinated CdS synthesis and optical readout.}
    \ac{sbl} executed 236 OT-2 commands to prepare a 96-well plate, followed by 19 robotic stages for transfer, sealing, re-grasping, and reader loading.
    Overlaid video frames and teal arrows illustrate plate handling.
    The heatmap shows fluorescence intensity at 560~nm after 8~h, with well H8 highlighted.
    The adjacent plots show the evolution of the H8 emission spectrum and its fluorescence intensity at 560~nm alongside the temperature trace.}
\label{fig:experiments_workflow}
\end{figure}

\Cref{fig:experiments_workflow} demonstrates the quantum dot synthesis experiment, which combines Opentrons liquid handling, plate sealing, and photophysics measurements.
The experiment highlights the challenges of end-to-end task planning and long-horizon robotic execution.
A rail-mounted robotic arm transfers a 96-well plate between instruments, requiring millimetre-level precision during pick-up and insertion while keeping the plate level throughout each trajectory.
\ac{sbl} autonomously performs these tasks after a human operator provides the initial information and approves the proposed plan.

\ac{sbl} used read-only API calls to establish the deck layout, pipettes, tip inventory, and sealer and reader capabilities, then adapted the requested chemistry to these constraints.
It wrote and verified a standalone Opentrons Python protocol offline, ran it in an Opentrons simulator matching the robot's software version, and froze the reviewed file by recording its SHA-256 hash.
The protocol consists of 236 commands covering setup, flow rates, tip, and liquid commands, all with timestamped event markers.
These actions were validated against the gateway's live OpenAPI schema, with the gateway configured to reject any change to the protocol's hash.
Before execution, \ac{sbl} re-read the protocol and required the equipment to be ready and idle.
It then sent the actions sequentially, recording device state before each request and confirming afterwards that no motion was in progress.
For the subsequent plate transfers, \ac{sbl} used the digital twin for motion planning, and visual and sensory feedback to guide grasping and placement.
The robotic arm first transferred the plate from the OT-2 to the PlateLoc for sealing.
To prepare the plate for insertion into the Cytation 5, \ac{sbl} then used an intermediate plate holder as a transfer station, where the arm re-grasped the plate along its short side.
The arm placed the plate in the reader's plate slot, and \ac{sbl} registered it with the Cytation 5 for optical readout.

The experiment synthesizes water-soluble end-capped cadmium sulfide quantum dots (CdS QDs), exploring several factors simultaneously: the reaction pH and the ratios of sulfide and stabilizing ligand with respect to Cd concentration. The reactions were performed in a 96-well microplate and incubated at 60 $^\circ\mathrm{C}$ in a microplate reader, where absorbance and photoluminescence measurements monitored how these factors affect nanoparticle formation and emission properties. \ce{Cd(NO3)2} and thiourea served as the cadmium and sulfide precursors. 3-mercaptopropionic acid (MPA) acted as the stabilizing ligand, coordinating to the nanocrystal surface through its thiolate group to restrict growth and aggregation \cite{Aboulaich2012,BelHajMohamed2018}. Alkaline buffers control the deprotonation state of MPA and, in turn, its coordination with \ce{Cd^2+}. At a fixed Cd concentration, the screen varied the Cd:MPA molar ratio (1:0--1:2), the Cd:thiourea molar ratio (1:1--1:2) and the initial buffer pH (8.5--10.0). 

After adding all reagents, the solution remains clear and colourless at room temperature, with homogeneous optical properties across the plate and almost no emission within 450--680~nm. Reaction monitoring over 8 h indicates no significant formation of amorphous precipitates, while a region with increasingly strong emission becomes notable in wells G7--H9. In well H8, for example, the increase in emission intensity above 500~nm and the shift of the emission peak from 520~nm to 560~nm are indicative of the progressive growth of MPA-capped CdS QDs \cite{Aboulaich2012}. The result revealed the optimal ligand loading of 0.75--1.5~eq, and the preferential QD formation at high pH values, which can be attributed to more advantageous release of sulfide via thiourea hydrolysis. It is consistent with the fact that ligands are required to stabilize soluble QDs after nucleation, yet excess ligands will inhibit QD growth.

\begin{figure}[!ht]
    \centering
    \includegraphics[width=\linewidth]{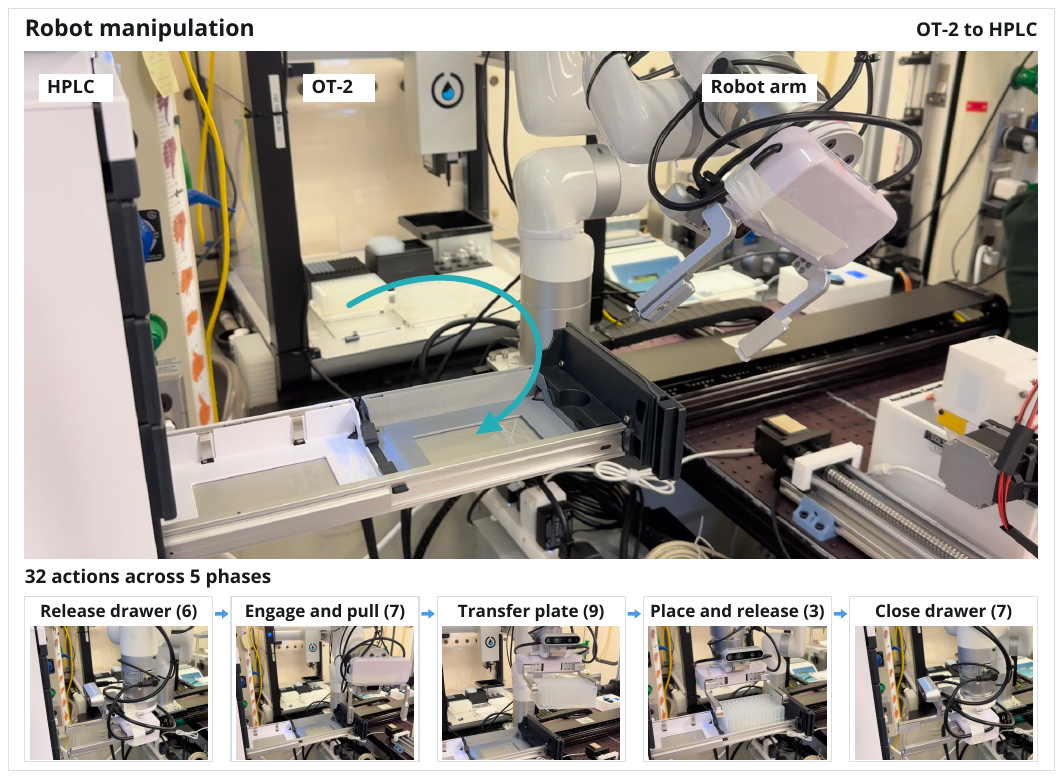}
    \caption{\textbf{Robot-assisted plate loading into an \ac{hplc}.} Teal arrows indicate the plate-transfer direction. Selected video frames illustrate five workflow phases. The 32-action workflow comprises 28 arm motions and 4 gripper actions.}
    \label{fig:hplc}
\end{figure}

\ac{sbl} also combined the previously developed \ac{hplc}-manipulation skills with plate transport to load a deep well plate from the OT-2 into an \ac{hplc} for sample analysis. \Cref{fig:hplc} illustrates the resulting workflow, with successive motions planned around drawer-position changes and whether the plate was held or released. The robot carried out the sequence, integrating free-space transport, sustained contact with the drawer, and plate release within the holder's narrow clearance. This example showed how instrument-specific skills could be composed into a workflow connecting liquid handling with sample loading for analysis.

\subsection{Building capabilities on an existing platform}
\label{sec:robochem}

\begin{figure}[p]
\centering
\includegraphics[width=\textwidth]{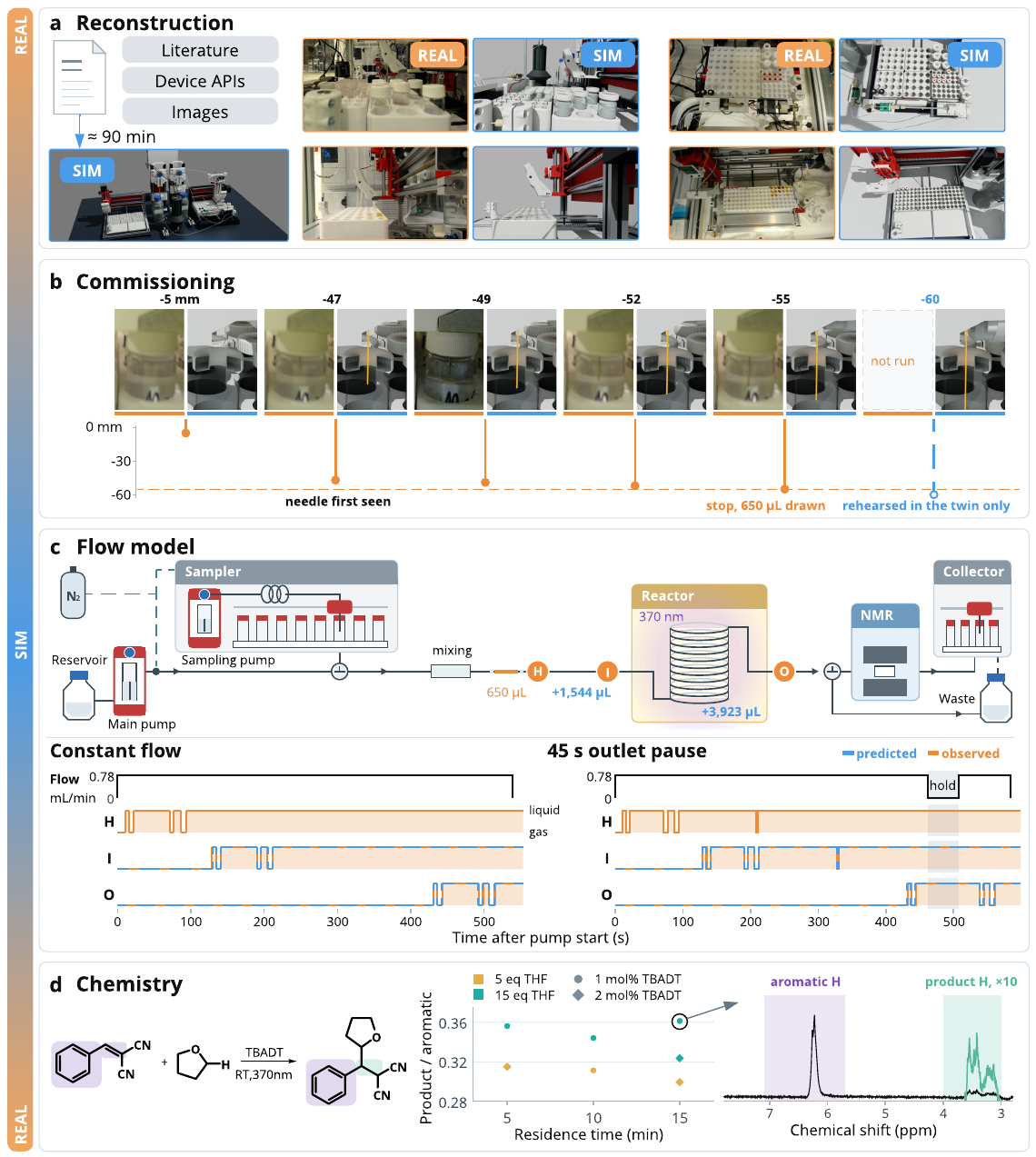}
\caption{\textbf{Building capabilities on RoboChem-Flex.}
(a) Digital-twin reconstruction from supplied resources, with four paired camera and rendered views.
(b) Needle descent into vial F-C3 as iteratively tried across simulation and real image observations at controller depths; the final position was only simulated as it approached the assumed vial floor.
(c) Digital twin internal flow path model with calibrated (blue numbers) and measured (orange numbers) inter-sensor distances and observed (orange) and predicted (blue, dashed) phase signals, with high and low signals denoting liquid and gas.
Predictions propagate boundaries detected at H using inter-sensor displacements calibrated on five packets. 
For a 650\,$\mu$L acetonitrile packet at 0.78\,mL\,min$^{-1}$ (left), outlet arrival errors are within 1.3\,s.
An agent-designed 45\,s pause (right) gives an outlet liquid window of 94.5\,s versus 94.6\,s predicted.
Both tests use the lamp off; the late pulse is chase liquid.
(d) Reaction scheme, corrected product-to-aromatic integral ratios for seven eligible runs versus nominal residence time, and baseline-corrected $^{1}$H \ac{nmr} spectrum of annotated run.
Purple and green bands mark aromatic and product integration windows; the product region is expanded tenfold.
}
\label{fig:robochem}
\end{figure}

\Cref{fig:robochem} examined how \ac{sbl} could develop operational and predictive capabilities on an existing \ac{sdl}.
RoboChem-Flex combines liquid handling, irradiation and inline analysis in a modular flow-photochemistry platform~\cite{slattery2024robochem,pilon2026robochemflex}.
Starting from the assembled apparatus, its publication and supporting files, low-level device \acp{api}, photographs and videos, \ac{sbl} built a digital twin and developed a model of the 28 published fluidic channels in approximately 90~min (\ref{fig:robochem}a).
The fluidic model includes tracked routes, syringe inventories and the sequence of liquid and gas segments.
This model was iteratively developed with physical trials to predict fluid transport, thereby establishing reusable procedures for priming and transfers (\ref{fig:robochem}b).

The agent calibrated fluid transport from sensor measurements and used the model to calculate downstream arrival times for different pump programmes.
The same calibration was used to predict programmes beyond those used to fit the model.
Across twelve held-out transits lasting 5--10~min, the observed transit midpoints differed from the predictions by less than 1.4~s, on the same order of magnitude as the response delay from device \ac{api} (Supplementary Section~\ref{sec:si-robochem-transport}).
These tests used acetonitrile in liquid-filled and nitrogen-cleared paths with the lamp off.

The model also predicted liquid and gas arrivals for programmes that accelerated flow, paused near the outlet or slowed flow during passage (\ref{fig:robochem}c).
A programme with a 45\,s pause produced an outlet liquid window of 94.48\,s compared with 94.60\,s predicted.
These comparisons show how the transport model can evaluate pump programmes for coordinating reactor passage and downstream handling.

The agent combined the handling procedures into a ten-run photochemistry campaign.
The reaction involved adding \ac{thf} to benzylidenemalononitrile using \ac{tbadt} as a photocatalyst~\cite{ravelli,laudadio}.
It passed reaction packets through the reactor under 370\,nm irradiation and positioned them for inline $^{1}$H \ac{nmr} acquisition (\ref{fig:robochem}d).
The campaign varied \ac{thf} loading, catalyst loading and nominal residence time.
The seven full-power measurements included in the comparison gave product-to-aromatic integral ratios of 0.30--0.36.

\subsection{Building scalable labs}
\label{building-labs}

One major challenge in laboratory automation is scalability, as existing systems are often tailored to specific hardware configurations and experimental workflows~\cite{bai2022platform,MacLeod2023howSDL}. SBL addresses this challenge by automatically designing, evaluating, and optimizing laboratory configurations in simulation before physical deployment. We demonstrate its scalability across increasingly complex setups, from single-arm and bimanual robotic systems to multi-station laboratories, where mobile robots coordinate material transfer between workstations. This enables SBL to adapt laboratory layouts, robotic configurations, and experimental workflows to different automation requirements.

\begin{figure}[htbp]
    \centering
    \includegraphics[width=\linewidth]{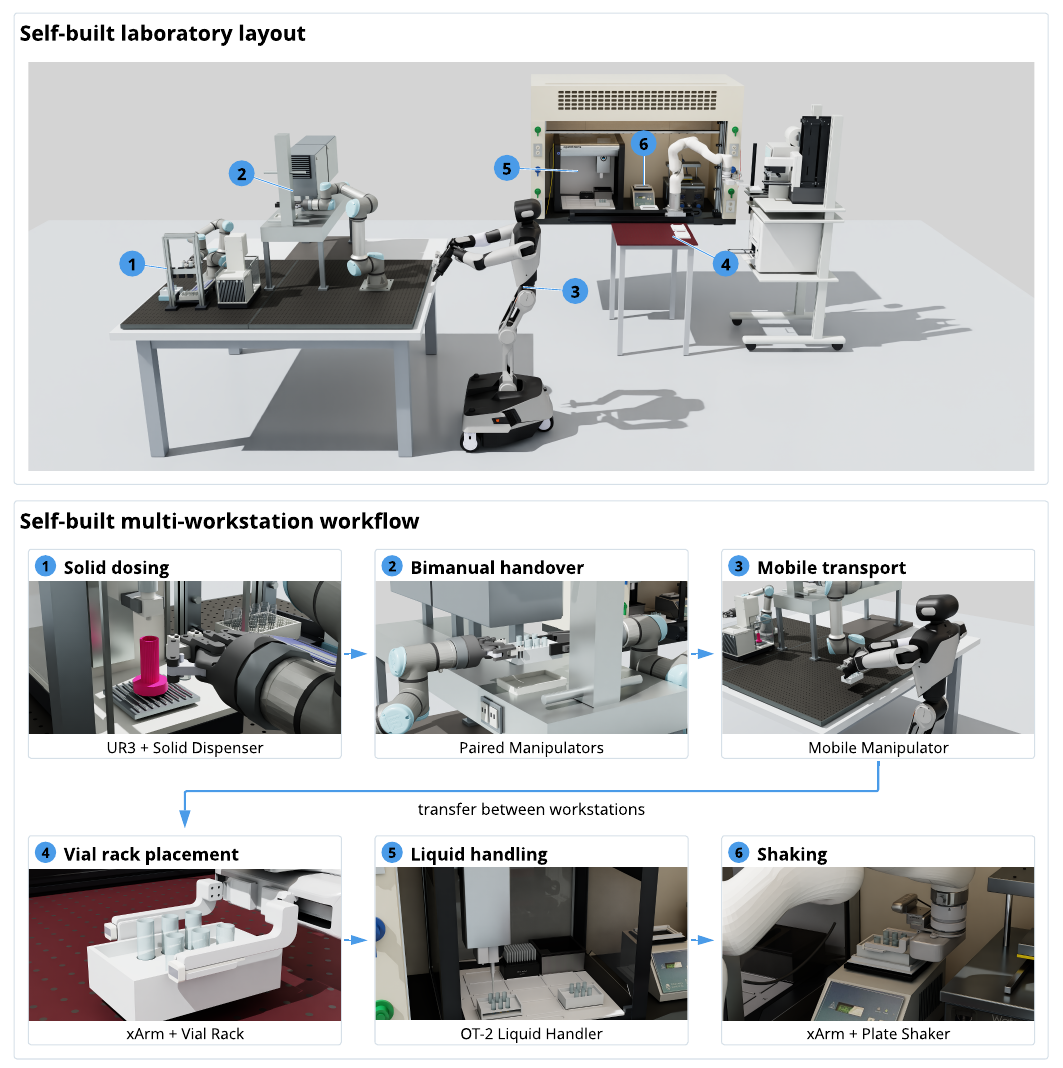}
    \caption{\textbf{Scaling self-building laboratories across robots and workstations.}
The reconstructed simulation layout (top) integrates a UR3/UR5 solid-handling workstation, a mobile manipulator, and an xArm-based liquid-handling workstation. Given a high-level buffer-preparation goal, SBL generates a coordinated workflow (bottom) in simulation and task-relevant powder and fluid simulation models for solid and liquid handling.}
    \label{fig:scalable_lab}
\end{figure}

\paragraph{\textbf{Scaling to different robots.}} We first demonstrate the adaptability of SBL to a solid-handling workstation, as shown in \Cref{fig:scalable_lab}. The task involves dispensing solid materials into six vials and transferring the vial rack to the opposite side of the workbench for subsequent experiments. SBL first reconstructs a digital twin from images of the physical laboratory, including particle-based simulation of solid dispensing. Because the user does not specify the vial and vial rack designs, SBL automatically generates CAD models that ensure secure vial placement and robotic graspability. A central shelf also prevents the smaller UR3 robot from directly transporting the vial rack across the workbench. To overcome this constraint, SBL autonomously plans a collision-free handoff between the UR3 and UR5 robots, enabling task completion through coordinated bimanual-independent robot manipulation.

\paragraph{\textbf{Scaling to multiple setups.}} Fixed-base robot arms are constrained by their limited workspace, while practical chemistry workflows often require devices distributed across multiple workstations. To demonstrate the scalability of SBL beyond a single robotic setup, we integrate the previously described UR-based solid-handling workstation with an xArm-based liquid-handling workstation to perform a buffer preparation workflow. Given only the high-level experimental goal, SBL autonomously generates a workflow that coordinates multiple robotic platforms and laboratory instruments. Specifically, SBL first leverages the UR workstation for solid dispensing, then plans and generates actions for a wheeled mobile manipulator to transport the vial rack between workstations. At the xArm workstation, the system coordinates placing the rack into an Opentrons liquid handler for reagent dispensing, followed by transfer to a shaker for mixing. This demonstration highlights the ability of SBL to extend existing robotic setups into an interconnected laboratory by automatically coordinating task allocation, inter-workstation transportation, and instrument operation.

\section{Discussion}

SBL brings experimental capabilities into the scientific process.
In the quantum-dot study, this meant adding missing controls, such as a joint-servoing interface for the arm, and establishing the conditions under which an operation succeeds, such as feedback-guided insertion when scene calibration was not precise enough.
The RoboChem-Flex study extends this process to operating procedures and a calibrated model for predicting fluid transport on an existing automated platform.
Both studies developed capabilities through physical interaction under human supervision; coordination across additional robots and workstations was examined in simulation.

What the agent can build depends on what the laboratory infrastructure lets it observe and verify~\cite{Guo2026stressTest}.
Some device failures produced no error from the \ac{api}, leaving the agent unable to tell success from failure~\cite{bar2012sila}.
Fine insertion became possible only after force and torque sensing exposed contact conditions that cameras and scene calibration could not~\cite{Zhou2025}.
On RoboChem-Flex, optical phase measurements made it possible to check transport predictions against observed fluid motion.

RoboChem-Flex shows how knowledge from physical trials can inform subsequent experimental planning.
A calibrated fluidic transport model allows candidate pump programmes to be evaluated before execution, with subsequent measurements testing the predictions and informing model refinement.
The separate exploratory flow model could support comparisons of equipment configurations by explicitly representing tube geometry and gas dynamics, subject to calibration and physical validation.

The digital twin also supports safety.
Simulation supports trajectory planning, collision checks, and exploration of hazardous scenarios before physical execution.
The resulting constraints inform workflow planning and runtime guardrails, including robot speed limits and force-triggered protective stops.
These guardrails must not rely on the model refusing unsafe instructions: tests on real robots show that GPT-6-Astra and other models will follow them~\cite{Sun2026RoboHarm}.
The validity of twin-based checks depends on the geometry, device states, and physical interactions represented in the twin~\cite{Ferrari2024digitaltwins}.
Collision checking covers modelled geometry; chemical hazards and uncertain contact conditions, such as the millimetre-scale grasping and placement in our workflow, require additional checks during physical execution.
Safe capability development therefore requires both space for exploration and enforced limits on physical actions, with execution observations used to refine the twin and its constraints~\cite{leong2025steering,Chen2026autonomyHarness}.

Human involvement remains necessary as new operating conditions emerge during long-horizon tasks~\cite{Yoshikawa2023llm_chem_robots,bai2024dynamic}.
Agents retrieve documentation, reconcile information, and execute code at a pace that makes continuous human inspection impractical.
During development, we observed both actions based on assumptions that required clarification and overly cautious motion decisions that obstructed task progress.
Disagreements between agents also produced repeated exchanges that required human intervention to resolve.
These experiences identify a communication requirement: the system must surface the assumptions, conflicting evidence, and operational consequences relevant to a decision.
Researchers retain responsibility for scientific objectives and supply physical facts that the available interfaces cannot observe.
On RoboChem-Flex, operator guidance helped establish a needle working depth that camera observations alone could not determine.
Twin-based checks and runtime guardrails provide a basis for delegating routine verification, while unresolved conditions bring the researcher enough context for intervention.
This division directs human expertise towards the unresolved decisions that limit further capability development.
This need for expert involvement, in turn, calls for training researchers to work in SBL environments, where human–agent–robot interaction becomes central, in contrast to both traditional human-led discovery and established SDLs running predefined workflows.

For this development process to accumulate capability, successful exploration must produce procedures that can be reused~\cite{llm_as_tool_maker,Huang2025CASCADE}.
Because agentic cycles were too slow for the chemical workflow, successful explorations were distilled into executable policies that run without the language model and return control to the agent when conditions fall outside their scope.
Encoding these conditions makes operational knowledge reusable beyond the original exploration.
Reuse across changed equipment or layouts requires checking those conditions and validating the adapted procedure~\cite{Rauschen2024universal}.
Related work tests reusable skills with visual feedback in simulation and on real robots~\cite{He2026Astra}, and programs that check their own outcomes in simulation~\cite{Si2026PyRUALean}.
Capability development thus becomes part of the scientific process rather than a prerequisite.
The next step is to measure whether successive cycles improve task performance and reduce the engineering effort and human intervention needed for new tasks, while retaining validated capabilities.
This would provide a quantitative basis for assessing self-improvement in physical \ac{ai}.

A global community of researchers can extend this process across laboratories by sharing the resources needed to develop and assess new capabilities~\cite{StriethKalthoff2023delocalized,leong2022object,bai2024dynamic,stach2021autonomous,Canty2025}.
Shareable resources include equipment models and digital-twin assets, device interfaces, operational skills, execution records, and safety guidelines, along with their applicability and validation conditions.
Shared safety standards would provide a common basis for assessing these resources, while deployment in another laboratory would still require local checks~\cite{Chen2026autonomyHarness}.
A global capability map of \acp{sdl} would connect scientific objectives to available instruments, skills, and resources, and expose the gaps that further capability development must address~\cite{bai2024dynamic,StriethKalthoff2023delocalized}.
Such a map could support planning investigations that draw on capabilities distributed across laboratories.
The result is a dynamic ecosystem for \acp{sbl}, in which scientific infrastructure evolves alongside scientific inquiry.
The case for self-building laboratories is strongest where engineers cannot follow the equipment: remote field stations, contaminated or hazardous sites, and even laboratories in space, which must operate under constraints on resources, communication, and crew time and develop new capabilities from what is already on board~\cite{sanders2023biological,Zhu2023Martian}.

\section{Methods}

\paragraph{\textbf{Software}}
We used \texttt{openai-gpt6-astra} to construct and modify laboratory representations, generate experimental workflows, and invoke planning and equipment-control tools. The digital twin was implemented in NVIDIA Isaac Sim 6.1 using \ac{usd}~\cite{nvidiaIsaacSim}. Equipment representations were assembled from available \ac{cad} models, reconstructed geometry, laboratory photographs, and device documentation. Articulated components were represented using joints, while liquid-handling states tracked liquid volume and composition. Robot motion planning used cuRobo~\cite{curobo_icra23} with robot collision-sphere representations and environment obstacles represented by meshes and cuboids. Collision checking was performed within the GPU planning pipeline. During workflow execution in the physical laboratory, all commands were issued through equipment-specific APIs, and robot and instrument status were recorded during operation.

\paragraph{\textbf{Acceleration Consortium SDL2 equipment}}
The experimental platform comprised a five-degree-of-freedom xArm5 robotic manipulator mounted on a motorized linear rail and equipped with a parallel gripper (BIO GRIPPER G2) and a 6-axis force/torque sensor for microplate handling (UFACTORY). The rail extended the robot's access across the liquid-handling, sealing, and measurement stations. Liquid dispensing was performed using an Opentrons OT-2 pipetting robot equipped with an eight-channel pipette (P300 Gen2) on the left, and a single-channel pipette (P1000 Gen2) on the right. The microplates were sealed using an Agilent PlateLoc Thermal Microplate Sealer and characterized using a BioTek Cytation5 multimode reader. We used 19-mm-tall 96-well flat-bottom (Agilent 201242-100) and 45-mm-tall deep 96-well U-shaped bottom (Agilent 5043-9302) polypropylene microplates. Intel RealSense D435i and D405 RGB-D cameras provided laboratory observations and close-range views of manipulation interfaces. Each instrument was exposed as an authenticated HTTP service with a read-only \verb|GET /status| endpoint, an OpenAPI schema, and documentation of available functions and valid input ranges. Measurement responses were saved incrementally as CSV files, together with run metadata. All machines connect through a private Tailscale network, with REST endpoints provided for instrument access. The central API checks the agent's authorization and records each device claim.

\paragraph{\textbf{Quantum-dot synthesis}}
Chemicals were purchased from Sigma-Aldrich and used as-is without further purification.
A series of pH 8.5, 9, 9.5 and 10 buffer solutions at 1~M concentration were prepared from 1~M \ce{NH4Cl} and 1~M \ce{NH4OH} stock solutions with volumetric ratios of 17.0:3.0, 12.8:7.2, 7.2:12.8, 3.0:17.0 (mL:mL), respectively.
The 0.1~M \ce{Cd(NO3)2}, 3-mercaptopropionic acid (MPA) and thiourea solutions were prepared by dissolving 0.617~g \ce{Cd(NO3)2 * 4H2O}, 0.213~g MPA and 0.152~g thiourea in deionized \ce{H2O} and diluting to 20~mL, respectively.
All operations were performed in fume hoods, and appropriate PPE was used for manual stock preparation. Only the source plate is pre-defined: the stock solutions were transferred to corresponding wells of the 45 mm 96-well deep well plate using volumetric pipettes and placed on the OT-2 pipetting robot for downstream reactions.  Column 1 held buffers (1.5 mL): rows A–B pH 8.5, C–D 9.0, E–F 9.5, G–H 10.0. Column 2 held 0.1~M \ce{Cd(NO3)2}, column 3 0.1~M MPA and column 4 0.1~M thiourea. Column 5 held deionized water (1.5 mL). Working stocks of 0.01~M \ce{Cd(NO3)2}, MPA and thiourea were made on deck by 1:10 dilution with water; the sulfide solution was diluted immediately before use.
The absorption values were measured at 350, 400, and 450 nm wavelengths. The emission spectra were measured in the 450--680~nm range and 10~nm intervals using 360~nm excitation wavelength.

\paragraph{\textbf{RoboChem-Flex equipment}}
The experimental platform~\cite{pilon2026robochemflex} comprised a three-axis sampler with main and auxiliary needles, a three-axis collector, 10~mL and 1~mL syringe pumps, an injection port and sample loop, and a mixing chamber.
The flow path included an approximately 3.9\,mL perfluoroalkoxy-tubing photoreactor in a 370\,nm lamp housing, a back-pressure regulator, a nitrogen line and a waste/analysis valve.
Inline analysis used a benchtop $^{1}$H \ac{nmr} spectrometer (Spinsolve, 61.8\,MHz).
Five optical phase sensors monitored fluid transport, and four cameras were added for this study.
Equipment control used version 3 of the device \ac{api}, which exposed device primitives and observations but did not supply the deck map.
Specified operating limits were 1\,mL\,min$^{-1}$ for the pumps and 1000\,mm\,min$^{-1}$ for the gantries.

\paragraph{\textbf{Flow photochemistry}}
Stocks were benzylidenemalononitrile at 2.0\,M, \ac{tbadt} at 17\,mM and neat \ac{thf}, with acetonitrile as diluent and carrier.
Each 650\,$\mu$L packet contained substrate at a nominal 75\,mM.
Stocks were combined in a mixing vial with seven reciprocating 175\,$\mu$L sampler strokes and injected as a gas-capped packet.
Reaction conditions varied \ac{thf} loading, catalyst loading and nominal residence time.
Pump rates of 0.26, 0.39 and 0.78\,mL\,min$^{-1}$ corresponded to nominal reactor residence times of 15, 10 and 5\,min.
The eligible comparison set used full lamp power.
Each packet was positioned in the \ac{nmr} flow cell and measured using WET solvent suppression~\cite{wet} on acetonitrile, with 64 scans, 3.3\,s acquisition time and 20\,s repetition time.
The reaction readout was the ratio of the product-window integral (3.0--4.0\,ppm) to the aromatic-window integral (5.7--7.1\,ppm) on the instrument's unreferenced scale.
Each integral used a linear baseline fitted to flanking windows.

\section*{Acknowledgments}
J.B. acknowledges funding from the Eric and Wendy Schmidt AI in Science Postdoctoral Fellowship Program, a program by Schmidt Futures.
E.S. and T.N. thank the European Union Innovation Council for the funding of ReaCtor (101099405).
A.A.-G. thanks Anders~G.~Fr{\o}seth for his generous support. A.A.-G. also acknowledges the generous support of Natural Resources Canada and the Canada 150 Research Chairs program. 
This work was supported by the AI2050 program of Schmidt Sciences.
\acknowAC
\acknowDARPA

\clearpage


{
\small
\bibliography{references}
\bibliographystyle{assets/plainnat}
}


\clearpage

\appendix

\setcounter{figure}{0}
\setcounter{table}{0}
\renewcommand{\thefigure}{S\arabic{figure}}
\renewcommand{\thetable}{S\arabic{table}}
\renewcommand{\theHfigure}{supp.\arabic{figure}}
\renewcommand{\theHtable}{supp.\arabic{table}}

\section{\acs{hplc} drawer manipulation and plate loading}
\label{sec:si-hplc}

\subsection{Instrument interface}

\begin{figure}[!ht]
    \centering
    \includegraphics[page=1,width=0.8\linewidth,height=0.425\textheight,keepaspectratio,trim=8.5mm 24mm 8.5mm 8.5mm,clip]{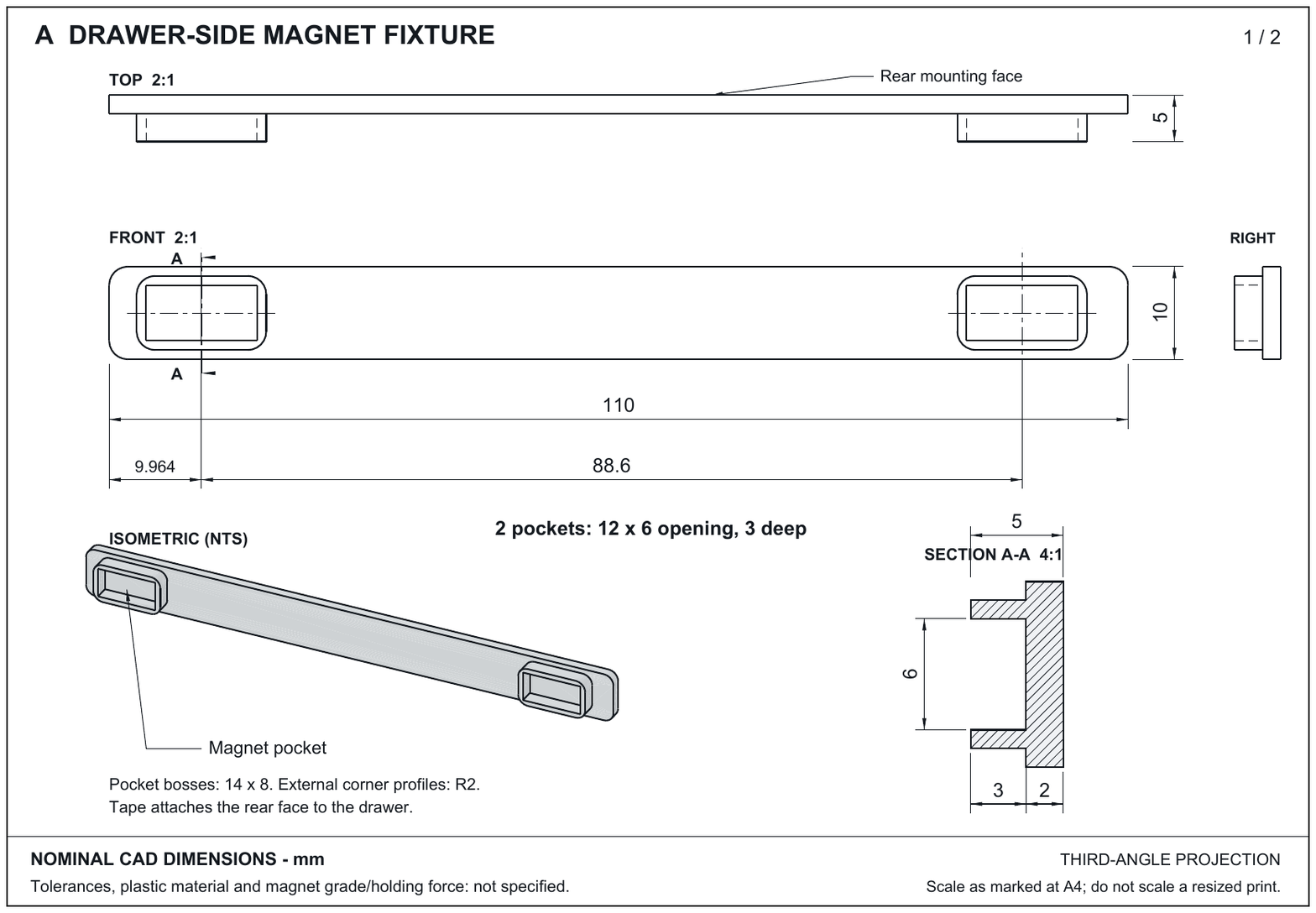}
    \par\vspace{2mm}
    \includegraphics[page=2,width=0.8\linewidth,height=0.425\textheight,keepaspectratio,trim=8.5mm 24mm 8.5mm 8.5mm,clip]{figs/si/si-hplc-magnetic-interface.pdf}
    \caption{\textbf{Magnetic fixtures for \acs{hplc} manipulation.}
    A: Drawer-side fixture.
    B: Left and right finger inserts.
    Dimensions are nominal values in millimetres and exclude mounting tape and magnets.}
    \label{fig:si-hplc-magnetic-interface}
\end{figure}

The sample-loading drawer couples a spring-loaded release mechanism to a sliding assembly carrying the plate holders.
We fitted the drawer front with a plastic fixture carrying two magnets before the manipulation trials to provide a magnetic interface with the gripper.
The curved inner surface of the front panel constrained where a gripper finger could bear against it during mechanical pulling.
The nominal \acs{cad} dimensions of the drawer fixture are $110 \times 10 \times 5$~mm; its two magnet pockets and those on the finger inserts each have a $12 \times 6$~mm opening and a depth of 3~mm (\Cref{fig:si-hplc-magnetic-interface}).

\subsection{Development of the manipulation strategy}

\begin{figure}[!ht]
    \centering
    \includegraphics[width=0.9\linewidth]{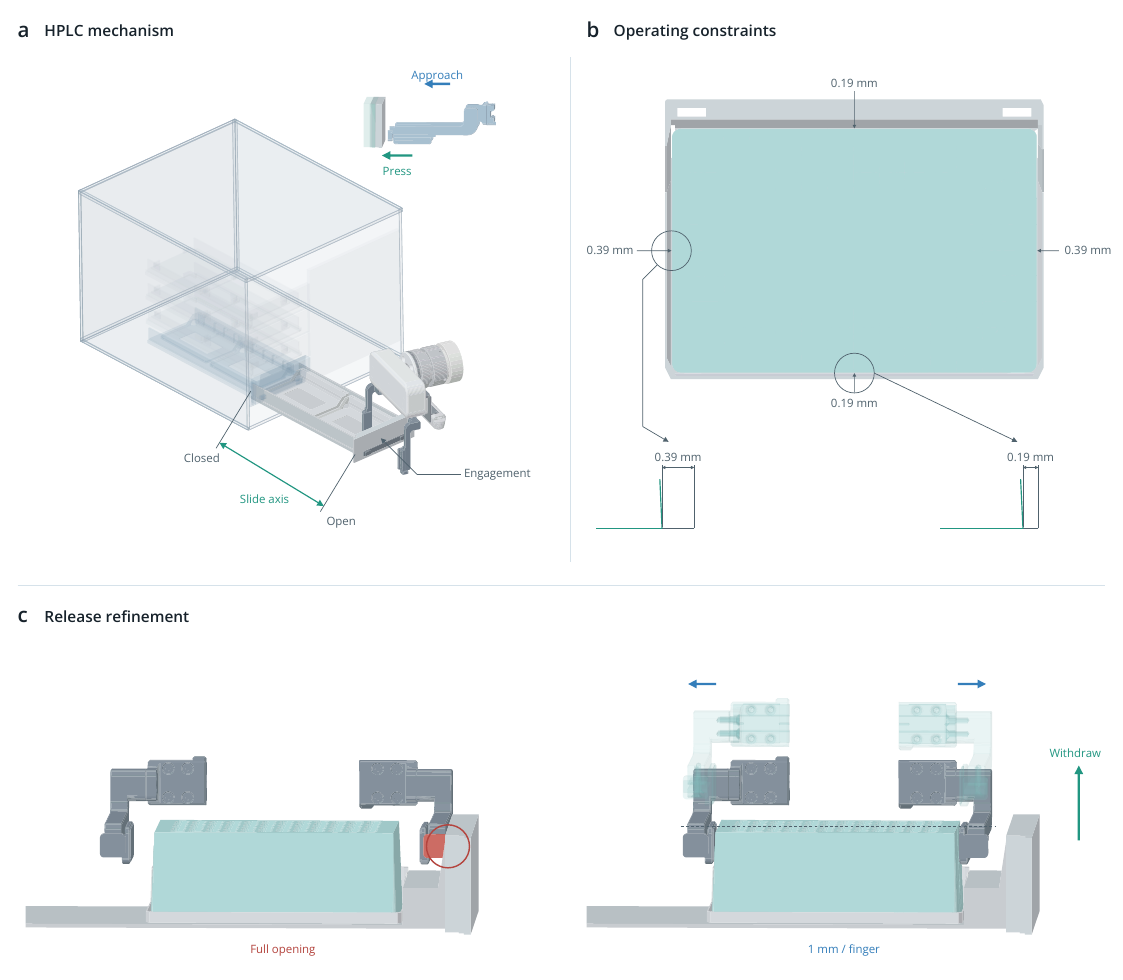}
    \caption{\textbf{Geometric constraints of \acs{hplc} manipulation.}
    \textbf{a} Drawer configurations with finger engagement behind the front panel to open the drawer along the slide axis. The inset shows approach and inward pressing to open the drawer.
    \textbf{b} Base view of the seated plate with nominal plate-holder \acs{cad} clearances.
    \textbf{c} Full gripper opening produces finger-panel interference (red), whereas 1~mm outward travel per finger permits release and withdrawal.}
    \label{fig:si-hplc}
\end{figure}

The digital twin represented the drawer front, magnetic counterparts and plate holders as a moving assembly, preserving their relative arrangement as the drawer moved.
In the seated configuration shown in \Cref{fig:si-hplc}b, the nominal plate-holder \acs{cad} clearances are approximately 0.39~mm and 0.19~mm per side along the long and short axes, respectively.
These clearances require submillimetre positioning during placement.
Gripper release was also constrained by the surrounding drawer geometry: full opening produced finger-panel interference in the digital twin, therefore requiring partial opening for release and vertical withdrawal (\Cref{fig:si-hplc}c).

The initial strategy used magnetic attachment to pull the drawer outward after pressing its release mechanism (8~mm inward).
During physical trials, the magnetic connection separated before the drawer had been fully opened, where recorded force/torque signals showed unloading during outward motion.
These observations motivated a transition to mechanical engagement for the remaining pull.
The agent explored candidate gripper orientations, insertion positions and pulling motions in the digital twin, refining the strategy and model using physical observations and sensor readings.
In the selected strategy, the gripper withdrew from the magnetic interface, reoriented with its fingers pointing downward, and inserted one finger behind the front panel while the other remained outside.
The contact location and insertion depth were constrained by the panel dimensions, the support beneath it and the clearance required by the complete gripper.

\subsection{Registration and drawer-state updates}

\begin{figure}[!ht]
    \centering
    \includegraphics[width=\linewidth]{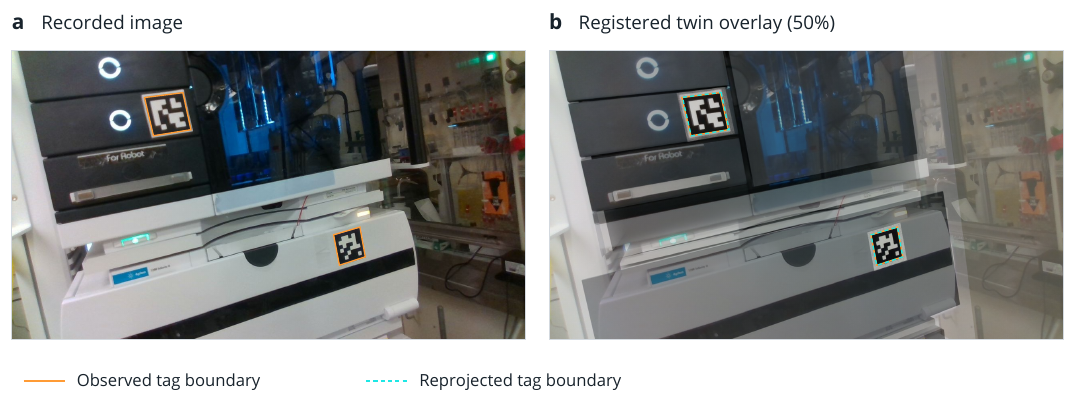}
    \caption{\textbf{AprilTag-based registration of the \acs{hplc} digital twin.}
    Captured image with AprilTags on the instrument and corresponding digital twin rendering overlaid on the image at 50\% opacity.
    Solid orange and dashed cyan outlines denote detected and reprojected tag boundaries, respectively.}
    \label{fig:si-hplc-apriltag-registration}
\end{figure}

AprilTag (\url{https://github.com/AprilRobotics/apriltag}) observations provided a spatial reference for registering the \ac{hplc} instrument in the laboratory's world frame (\Cref{fig:si-hplc-apriltag-registration}).
During development, observations of the closed and open drawer, together with operator-supplied measurements, were used to refine its position, panel orientation and opening geometry.
Comparing these configurations helped estimate drawer displacement and update the holder positions in the twin.

For trials with recorded unloading events, force/torque changes and synchronized robot joint reports also informed estimates of the release state.
These estimates depended on the assumption that the drawer followed the tool while engaged and stopped after detachment during slow withdrawal.
State updates were made at the available observation or review stages.

\subsection{Operation sequence and force/torque references}

The resulting workflow combined drawer opening, plate transfer, constrained release and drawer closure (\Cref{tab:hplc-sequence}).
Force/torque signals and robot joint reports were recorded together and compared by operation phase.
The traces captured changes during pressing, pulling, disengagement, plate placement and closure, providing empirical references for automatic detection of departures from previously observed responses.

\begin{table}[!ht]
    \centering\small
    \caption{Sequence of \acs{hplc} drawer manipulation and plate loading.}
    \label{tab:hplc-sequence}
    \begin{tabular}{@{}>{\raggedright\arraybackslash}p{0.16\linewidth}>{\raggedright\arraybackslash}p{\dimexpr0.84\linewidth-2\tabcolsep\relax}@{}}
        \toprule
        Phase & Action and relevant state \\
        \midrule
        Release drawer & Approach the front panel, press inward and withdraw from compression. \\
        Engage and pull & Disengage the magnetic interface, reorient, insert one finger behind the panel and pull along the drawer axis. \\
        Transfer plate & Leave the drawer, grasp the deep-well plate on the liquid handler and transfer it above the holder at the open drawer position. \\
        Place and release & Lower the plate into the holder, partially open the gripper and withdraw upward within the available clearance. \\
        Close drawer & Reposition the empty gripper, push the drawer inward, press the release mechanism and retreat. \\
        \bottomrule
    \end{tabular}
\end{table}

\newpage

\section{RoboChem-Flex flowchemistry}
\label{sec:si-robochem}

\subsection{Information access}
\label{sec:si-robochem-information}

The agent was provided with the RoboChem-Flex publication and supporting information~\cite{pilon2026robochemflex}, CAD, photographs and videos, four camera feeds, and low-level device \ac{api}. 
The supporting information of the paper details the 28 fluidic channels that track routes, syringe inventories and liquid/gas segments.
The existing orchestration software, deck map and high-level experimental workflows were withheld. The \ac{api} exposed motion, pumping, valves, illumination and instrument observations. Specified limits were 1\,mL\,min$^{-1}$ for pumps and 1000\,mm\,min$^{-1}$ for gantries. Human operator supplied depth references, priming and source-inventory guidance.

\subsection{Reconstruction and commissioning}
\label{sec:si-robochem-commissioning}

Camera comparisons guided scene refinement. Eighteen reactor-span measurements covered 3.80--4.10\,mL, consistent with the initial 3.90--4.00\,mL representation. Iterative simulation and execution with camera checks were used to locate the visible tip, where the needle descent to $z=-55$\,mm. A nominal 650\,$\mu$L draw produced a 638--642\,$\mu$L optical packet width.

\ac{nmr} commissioning included acquisition-parameter selection and scan-guided slug positioning. The agent settled on approximately 3.3\,s acquisition time per scan. It alternated rapid two-scan spectra with small pumping increments to locate the slug in the flow cell, then held it stationary for a 64-scan acquisition.

\subsection{Calibrated transport prediction}
\label{sec:si-robochem-model}

Tests used acetonitrile with the lamp off, the \ac{nmr} disconnected and the outlet directed to waste. H, I and O denote the handler outlet, reactor inlet and reactor outlet sensors. Routes were either liquid-primed or nitrogen-cleared. Phase changes were bracketed by adjacent sensor readings, using request-time and pump-counter records or buffered acquisition at approximately 9\,Hz per channel.

Calibration used Q1/P2/P3/P4/P8, with 16 H--I and 20 I--O boundary events. Event displacement midpoints were averaged within each run, then run means were weighted equally. This gave $D_{HI}=1544.2555\,\mu\mathrm{L}$ and $D_{IO}=3922.559\,\mu\mathrm{L}$. A boundary observed at H at time $t_H$ reaches sensor $X$ at the first $t_X$,
\begin{equation}
\int_{t_H}^{t_X}q_{\mathrm{motor}}(t)\,dt=D_{HX},\qquad D_{HO}=D_{HI}+D_{IO},
\label{eq:robochem-arrival}
\end{equation}
where $q_{\mathrm{motor}}$ is pump-motor volume per unit time. At 0.78\,mL\,min$^{-1}$, the predicted reactor transit is 301.735\,s. Pump holds add to transit time; the model assumes no coast during a stop. Recipe forecasts used a preparation template; H-conditioned forecasts used the packet measured in the current run. Both were saved before the relevant execution or first predicted inlet arrival, respectively.

\paragraph{Transport comparisons.}
\label{sec:si-robochem-transport}

\Cref{tab:robochem-flow} details specific packets investigated in this comparison.
The transit comparison included sixteen calibration transits and twelve held-out transits from P9/P13/T2/T4. Without refitting, all twelve held-out midpoints were within 1.4\,s of prediction. All eight P9/P13 brackets lay within the frozen forecast envelopes; P9 included its measured hold under the prespecified rule. Multiple boundaries within a packet are correlated observations.

The main figure compares continuous flow (R1) with a pump pause (S2), using nominal 50/650/50\,$\mu$L gas/liquid/gas packets and 7\,mL main-pump displacement. R1 gave core-front/tail transits of 302.40/302.59\,s against 301.74\,s predicted, and an outlet liquid window of 49.50\,s against 49.32\,s. With a measured 45.20\,s hold, S2 gave an outlet window of 94.48\,s against 94.60\,s predicted from H. All fourteen associated downstream boundaries were within 19.2\,$\mu$L of prediction, meeting the chosen 75\,$\mu$L tolerance. Prediction and observation share the saved pump-start origin; no outlet-time alignment was fitted. S1/S3 displacement comparisons gave outlet residuals within 12/32.3\,$\mu$L; S1 used its measured split packet and S3 used retrospective boundary association.

\begin{table}[!htbp]
\centering\small
\caption{Transport experiments and their evidential roles. Rates are in mL\,min$^{-1}$; packet volumes and gas draws are commanded values. Unless stated otherwise, the nominal liquid packet was 650\,$\mu$L. Calibration rows refer to the final fit, even where an earlier forecast preceded the run.}
\label{tab:robochem-flow}
\begin{tabular}{@{}>{\raggedright\arraybackslash}p{0.08\linewidth}>{\raggedright\arraybackslash}p{0.33\linewidth}>{\raggedright\arraybackslash}p{\dimexpr0.59\linewidth-4\tabcolsep\relax}@{}}
\toprule
Run & Test & Role and result\\
\midrule
Q1 & 0.26; liquid-filled route & Calibration; initial rate-transfer envelope.\\
P2 & 0.78; liquid-filled route & Calibration; nominal gas draws 50/75\,$\mu$L.\\
P3, P4 & 0.78; larger rear/front gas draw & Calibration; respectively 50/150 and 150/75\,$\mu$L gas draws.\\
P8 & 0.39; liquid-filled route & Calibration at an intermediate rate.\\
P9 & 0.78 with a measured 50--53\,s stop & Held out; all four transit brackets inside the frozen envelope with the measured-hold rule.\\
P13 & 0.39; 325\,$\mu$L liquid packet & Held out; all four transit brackets inside the frozen envelope.\\
T2, T4 & 0.78; baseline/halved injection rate & Later held-out tests; four core-boundary transits of 302.22--302.47\,s versus 301.735\,s predicted.\\
P6, P7 & 0.78/0.26; nitrogen-cleared route & Diagnostic bulk selection: short pulses flagged retrospectively; 13 of 14 transit brackets inside, all 14 overlapping the earlier envelopes.\\
T1, T3 & No gas caps / doubled gas-draw rate & Preparation controls; T1 merged liquid with chase, while T3 gave a 643.6\,$\mu$L core at H. Not part of the twelve held-out transit points.\\
R1 & 7\,mL at 0.78; continuous flow & Later comparison: core-front/tail transits 302.40/302.59\,s versus 301.74\,s predicted; outlet window 49.50\,s versus 49.32\,s.\\
S1 & 3.5\,mL at 0.39, then 3.5\,mL at 0.78 & Displacement comparison: nine associated outlet boundaries agreed with the upstream-conditioned prediction within 12\,$\mu$L.\\
S2 & 6\,mL at 0.78; 45\,s hold; 1\,mL at 0.78 & Completed prespecified test; fourteen downstream boundaries within 19.2\,$\mu$L of the H-conditioned forecast.\\
S3 & 6\,mL at 0.78, then 1\,mL at 0.26 & Displacement comparison: nine retrospectively associated outlet boundaries agreed with prediction within 32.3\,$\mu$L.\\
\bottomrule
\end{tabular}
\end{table}

\subsection{Hydraulic model and choice of transport predictor}
\label{sec:si-robochem-gas}

The agent also built a one-dimensional hydraulic model with gas compression and flow resistance. Both models used measured H boundaries and recorded pump execution. They were calibrated on T2/T4 and evaluated without refitting on R1/S2, with two packets and eight core-boundary observations per group. The hydraulic comparison represented the displacement-derived volumes as horizontal 0.8\,mm-bore tubes and fitted two additional sensor offsets. Nominal assumptions were 0.37\,mPa\,s liquid viscosity, 100\,kPa initial gas pressure, zero surface tension and a 0.15\,mL outlet extension.

Both models gave approximately 1.81\,$\mu$L mean absolute boundary error on the evaluation packets. Their difference (errors in pump-motor coordinates) was much smaller than the observation brackets. The displacement model (\Cref{eq:robochem-arrival}) was retained because it supplied sufficient working predictions with fewer fitted quantities and no gas-pressure initialisation.

The hydraulic solver uses cumulative tube volume,
\begin{equation}
s(\ell)=\int_0^\ell\pi r(\xi)^2\,d\xi,
\label{eq:robochem-volume-coordinate}
\end{equation}
where $r$ is the internal radius. For gas cap $k$, interfaces $a_k,b_k$ enclose volume $V_k=b_k-a_k$ at absolute pressure $p_k$. Adjacent liquid flows $q_{k-1},q_k$ give
\begin{equation}
\dot a_k=q_{k-1},\quad \dot b_k=q_k,\quad
\dot V_k=q_k-q_{k-1},\quad p_kV_k=p_{k,0}V_{k,0}.
\label{eq:robochem-gas}
\end{equation}
The pump sets $q_0(t)$. Each incompressible liquid plug obeys
\begin{equation}
R_jq_j=p^{\ell}_{j,\mathrm{in}}-p^{\ell}_{j,\mathrm{out}}-\rho g\Delta z_j,
\qquad R_j=\frac{8\mu}{\pi}\int_{\ell_{j,\mathrm{in}}}^{\ell_{j,\mathrm{out}}}\frac{d\ell}{r(\ell)^4},
\label{eq:robochem-liquid}
\end{equation}
with viscosity $\mu$, density $\rho$ and elevation change $\Delta z_j$. The optional capillary closure is $p^{\ell}=p_k-2\sigma\cos\theta/r$, with surface tension $\sigma$ and fixed contact angle $\theta$. An exposed gas region of volume $G$ vents according to
\begin{equation}
M=pG,\qquad \dot M=-J,\qquad J=\frac{p^2-p_{\mathrm{atm}}^2}{2R_g},\qquad q_{g,\mathrm{atm}}=J/p_{\mathrm{atm}},
\label{eq:robochem-vent}
\end{equation}
where $R_g$ is the gas-region Poiseuille resistance. Unlike displacement transport, this model permits pressure relaxation and liquid motion during a pump hold. It remains available for exploring tube dimensions, reactor volumes and pump schedules. Predictions depend on assumed geometry and fluid properties; films, breakup, contact-angle hysteresis, multidimensional mixing and irradiation fields are not resolved.

\subsection{Photochemistry and spectral processing}
\label{sec:si-robochem-photochemistry}

\begin{table}[!ht]
\centering\small
\caption{Reaction conditions and corrected spectral ratios. THF is in equivalents, TBADT in mol\%, $t$ in minutes and lamp setting in percent. Each condition was run once. O1--O7 form the main comparison.}
\label{tab:robochem-chemistry}
\begin{tabular}{@{}lrrrrr@{}}
\toprule
Run & THF & TBADT & $t$ & Lamp & Ratio\\
\midrule
Initial & 5 & 1 & 10 & 50 & 0.3068\\
O1 & 5 & 1 & 10 & 100 & 0.3109\\
O2 & 15 & 2 & 15 & 100 & 0.3236\\
O3 & 15 & 1 & 10 & 100 & 0.3438\\
O4 & 5 & 2 & 15 & 100 & 0.2987\\
O5 & 5 & 2 & 5 & 100 & 0.3143\\
O6 & 15 & 1 & 15 & 100 & 0.3604\\
O7 & 15 & 1 & 5 & 100 & 0.3557\\
O8 & 15 & 2 & 5 & 100 & ---\\
O9 & 15 & 1 & 15 & 100 & 0.3052\\
X & 11 & 1.5 & 5 & 75 & 0.3318\\
\bottomrule
\end{tabular}
\end{table}

All chemicals were sourced from Merck KGaA and used without further purification. Each nominal 650\,$\mu$L packet contained 48.75\,$\mu$mol benzylidenemalononitrile (75\,mM), neat \ac{thf} and \ac{tbadt} in acetonitrile. Stock draws were 24.375\,$\mu$L of 2.0\,M substrate, $4.0625e$\,$\mu$L \ac{thf} for $e$ equivalents and $28.676p$\,$\mu$L of 17\,mM catalyst for $p$\,mol\%, with acetonitrile to volume. Seven reciprocating 175\,$\mu$L sampler strokes preceded gas-capped injection. Flow rates of 0.26/0.39/0.78\,mL\,min$^{-1}$ gave nominal 15/10/5\,min residence times in the approximately 3.9\,mL reactor under 370\,nm illumination.

\begin{figure}[p]
\centering
\includegraphics[width=\linewidth]{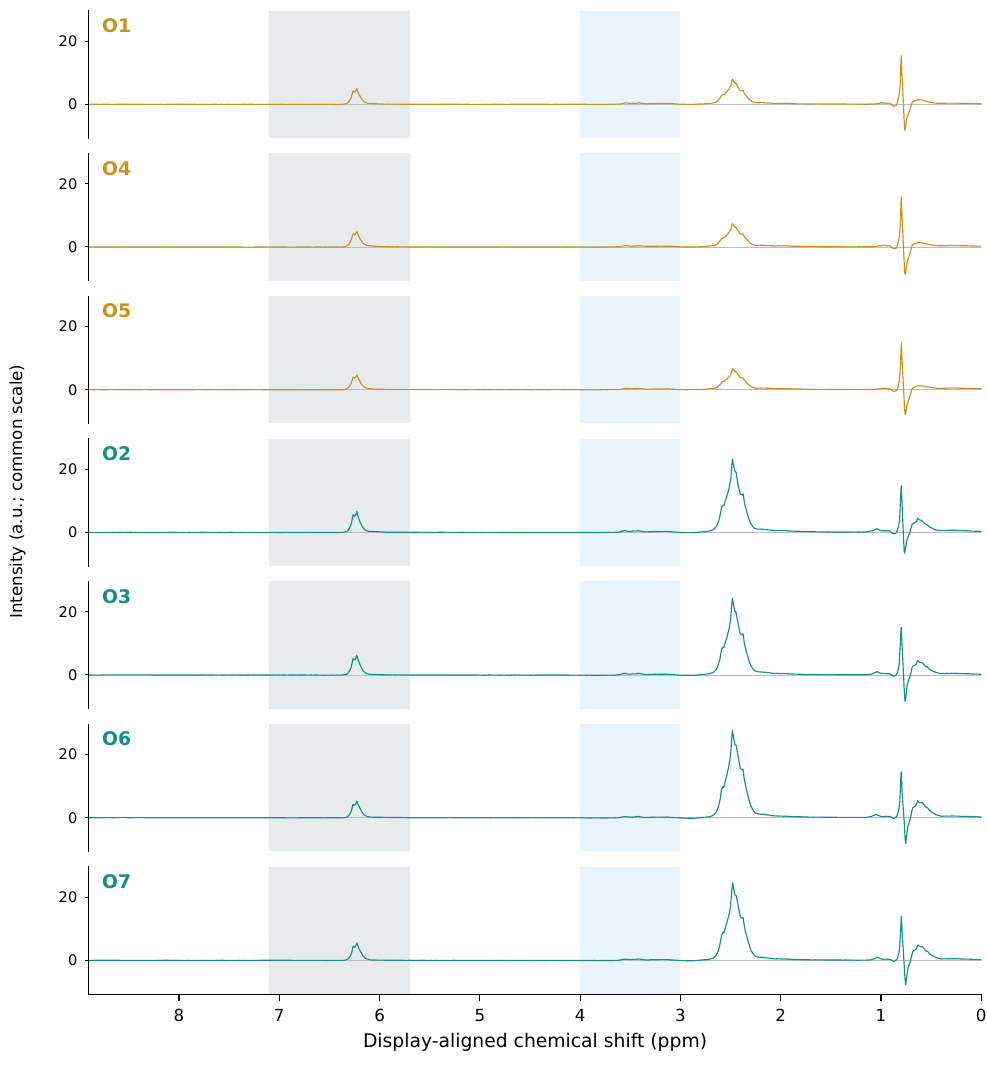}
\caption{Broader $^1$H \ac{nmr} spectra of O1--O7. Instrument-processed exports are display-aligned on a common intensity scale, without additional baseline correction, per-trace normalisation or vertical clipping. Gold/teal indicate 5/15 equivalents of THF. Grey/blue bands mark aromatic/product integration regions.}
\label{fig:robochem-si-spectra}
\end{figure}

After scan-guided positioning, 64-scan $^1$H spectra were acquired at 61.8\,MHz with WET acetonitrile suppression and $^{13}$C decoupling (\Cref{fig:robochem-si-spectra}). Acquisition time was 3.3\,s, repetition time 20\,s and receiver gain 25; 8192 points were zero-filled to 32768. Each acquisition lasted approximately 1301\,s. \Cref{tab:robochem-chemistry} gives the conditions and spectral ratios.
O8 failed preparation and supplied no 64-scan spectrum. The fragmented O9 packet was excluded before scoring. X was a separate condition check; a substrate reference used 650\,$\mu$L of 2.0\,M stock through the dark reactor.
The score, specified before O1, was the signed product-window integral divided by the aromatic-window integral after linear baseline correction. Native-scale windows were 3.0--4.0 and 5.7--7.1\,ppm; baseline flanks were 2.85--2.98/4.10--4.30 and 5.30--5.55/7.30--7.55\,ppm, respectively. Windows followed each run's acetonitrile offset relative to 0.802\,ppm. Trapezoidal integration retained negative values and reproduced archived scores within 0.2\,\%.

\end{document}